\documentclass{article}

\usepackage[preprint]{neurips_2025}

\usepackage[utf8]{inputenc} 
\usepackage[T1]{fontenc}    
\usepackage{hyperref}       
\usepackage{url}            
\usepackage{booktabs}       
\usepackage{amsfonts}       
\usepackage{nicefrac}       
\usepackage{microtype}      
\usepackage{xcolor}         

\usepackage{graphicx}
\usepackage{subcaption}

\usepackage{amsmath}
\usepackage{amssymb}
\usepackage{mathtools}
\usepackage{amsthm}

\usepackage{enumitem}
\usepackage[utf8]{inputenc} 
\usepackage[T1]{fontenc}    
\usepackage{hyperref}       
\usepackage{url}            
\usepackage{amsfonts}       
\usepackage{nicefrac}       
\usepackage{xcolor}         
\usepackage{comment} 
\usepackage{tcolorbox}
\usepackage{wrapfig}
\usepackage{adjustbox} 
\usepackage{multirow}
\usepackage{listings}
\usepackage{mdframed}
\usepackage{fancyvrb}
\usepackage{caption}
\usepackage{fvextra}           
\usepackage{newfloat}

\DeclareFloatingEnvironment[
  fileext=lol,
  listname={List of Listings},
  name=Listing
]{listing}

\usepackage[capitalize,noabbrev]{cleveref}
\theoremstyle{plain}

\theoremstyle{definition}

\theoremstyle{remark}

\usepackage[textsize=tiny]{todonotes}

\title{Correct Reasoning Paths Visit Shared Decision Pivots}

\author{%
\textbf{Dongkyu Cho$^{1}$\thanks{Paper written during Dongkyu's internship at Amazon}, \quad Amy B.\,Z. Zhang$^{2}$, \quad Bilel Fehri$^{2}$, \quad Sheng Wang$^{2,3}$}\\
\textbf{Rumi Chunara$^{1}$, \quad Hengrui Cai$^{2,4}$, \quad Rui Song$^{2}$\thanks{Corresponding Author}}\\[0.3em]
$^{1}$ New York University \quad $^{2}$ Amazon \quad $^{3}$ University of Washington \quad $^{4}$ UC Irvine\\[0.3em]
\textit{dongkyu.cho@nyu.edu, \{amybzz, bffehri\}@amazon.com, swang@cs.washington.edu,}\\
\textit{rumi.chunara@nyu.edu, hengrc1@uci.edu, ruisong@amazon.com}
}

\begin{document}

\maketitle

\begin{abstract}
Chain‑of‑thought (CoT) reasoning exposes the intermediate thinking process of large language models (LLMs), yet verifying those traces at scale remains unsolved. In response, we introduce the idea of decision pivots—minimal, verifiable checkpoints that any correct reasoning path must visit. We hypothesize that correct reasoning, though stylistically diverse, converges on the same pivot set; incorrect ones violate at least one pivot. Leveraging this property, we propose a self‑training pipeline that (i) samples diverse reasoning paths and mines shared decision pivots, (ii) compresses each trace into pivot‑focused short‑path reasoning using an auxiliary verifier, and (iii) post-trains the model using its self-generated outputs. The proposed method aligns reasoning without ground truth reasoning data or external metrics. Experiments on standard benchmarks such as LogiQA, MedQA, and MATH500 show the effectiveness of our method.
\end{abstract}

\section{Introduction}\label{sec:intro}

Large language models (LLMs) can now solve diverse problems when prompted to articulate their chain‑of‑thought (CoT) reasoning \citep{wei2022chainofthought}. Yet these reasoning paths are often redundant, self‑contradictory, or logically unsound, even when they culminate in the right answer. Current training pipelines such as reinforcement learning from human feedback (RLHF) \citep{ouyang2022training} evaluate entire responses, providing only coarse signals that do little to supervise the underlying reasoning process.  As a result, we still lack a scalable way to verify and train high‑quality explanations without resorting to expensive human annotation. We claim that many reasoning tasks are governed by decision pivots: minimal, semi-verifiable checkpoints (e.g., keyword, key logic) that every sound explanation must traverse.  Echoing the proverb “All roads lead to Rome”, correct CoTs succeed for the same reason: they visit the shared pivot set; whereas incorrect paths fail in many different ways, omitting or contradicting at least one pivot.  Exploiting this observation, we propose a self‑training method, \textsc{ROMA} (Reasoning Optimization via Multi-path Aggregation), that aligns a model’s reasoning without utilizing human-annotated rationales or handcrafted metrics.

Our method operates in three stages.  First, we draw $K$ diverse CoT reasoning paths for each input sample, bootstrapping the pivots shared by the successful reasoning paths \citep{zelikman2022starbootstrappingreasoningreasoning}.  Second, we merge multiple reasoning paths into a \textit{short‑path reasoning} (SPR) that focuses on the pivots and removes redundant steps.  Finally, we post-train the model with RLHF \citep{rafailov2023direct}, treating SPRs as preferred positives and individual reasoning paths as negatives. The entire loop is model‑driven and repeatable at scale \citep{shafayat2025largereasoningmodelsselftrain}. Experiments on \textsc{LogiQA}\citep{liu2020logiqa}, \textsc{MedQA}\citep{jin2021disease}, and \textsc{MATH500}\citep{hendrycks2021measuring} yield simultaneous gains in solving downstream task accuracy and in reasoning quality, while confirming the method’s generality across commonsense, expert (e.g., medical), and math domains.

\vspace{-2mm}
\paragraph{Contributions. }
\begin{itemize}[leftmargin=*, itemsep=0pt, parsep=0pt, topsep=0pt]
  \item We formalize \textit{decision pivots} and show that correct reasoning paths intersect on a compact set of pivots.
  \item We propose \textsc{ROMA}, a bootstrap$\rightarrow$rewrite framework that aggregates $K$ paths and, via a fine-tuned verifier, distills a compact, pivot-focused short-path reasoning (SPR).
  \item We build a self-training pipeline that leverages SPR to improve both task accuracy and reasoning quality across benchmarks (\textsc{LogiQA}, \textsc{MedQA}, \textsc{MATH500}).
  \item We introduce a decision pivot-based reasoning metric and show it complements general CoT metrics (e.g., ROSCOE) for preference-data selection.
\end{itemize}

\section{Preliminaries \& Positioning}\label{sec:related}
\begin{figure}[ht]
    \centering
\includegraphics[width=0.9\textwidth]{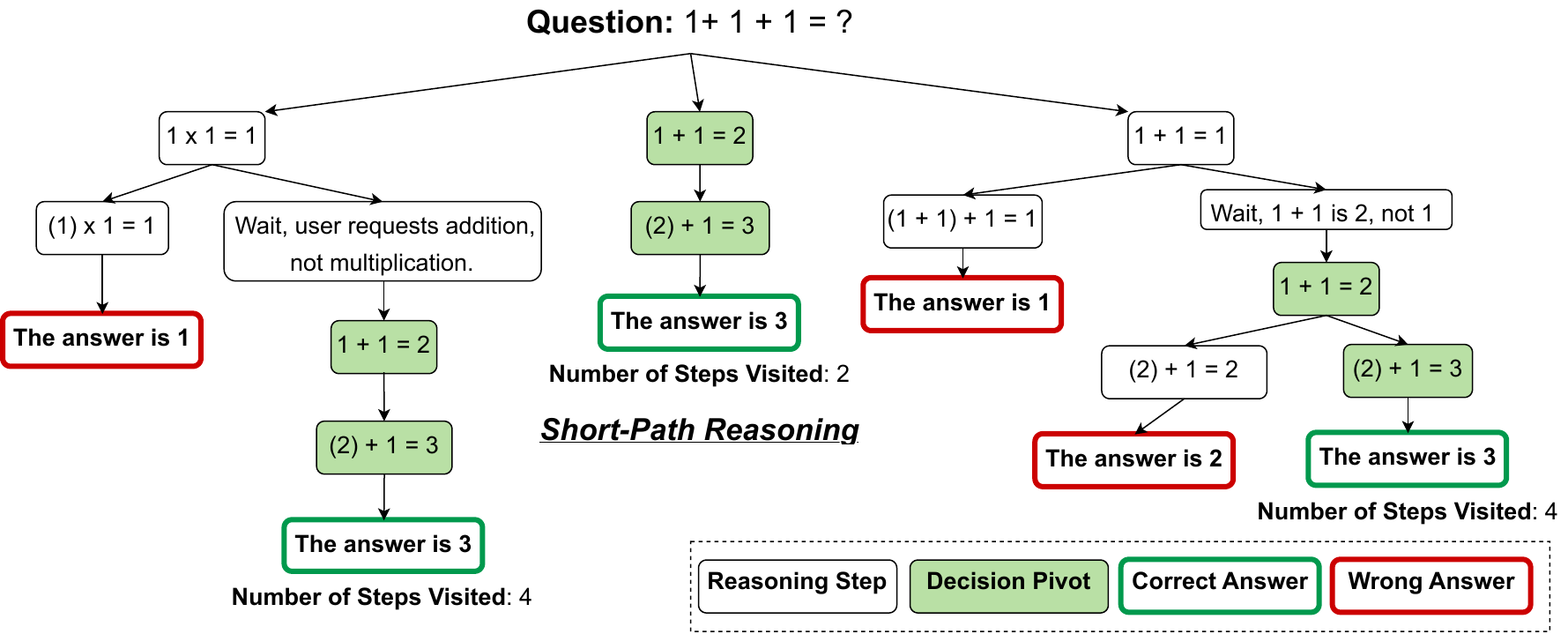}
    \caption{A model's chain-of-thought reasoning may take various paths to reach a final decision, naturally visiting redundant or incorrect thought steps. However, such errors are not easily captured by existing methods. Instead, we introduce the concept of decision pivots, which are a set of key information that a model's reasoning path must visit to reach a certain decision. In this sense, we aim to generate a concise, factual reasoning path that focuses on the decision-pivots, which we denote as Short-Path Reasoning (SPR).} 
    \label{fig:decision-pivots}
\vspace{-3mm}
\end{figure}

\paragraph{Post-training. }
 Post-training adapts a pre-trained model to produce human-aligned outputs. Typically, this involves supervised fine-tuning (SFT) on curated human-aligned data, followed by reinforcement learning (RL) \cite{ouyang2022training} on preference-labeled examples. A key challenge is constructing reliable preference data. One line of work derives labels from explicit metrics \cite{golovneva2023roscoesuitemetricsscoring,prasad2023recevalevaluatingreasoningchains,golovneva2023pathfinder} or from model-based feedback \cite{liu-etal-2023-g,kim2023prometheus,bai2022constitutional,lee2024rlaif} (LLM-as-judge). RLVR (reinforcement learning with verifiable rewards) removes human preference signals entirely by using programmatic verifiers to assign binary rewards, which are effective in domains like math or coding, where correctness is objectively checkable \cite{deepseek2024r1zero}. Extensions seek to broaden this to more open ended reasoning \cite{ma2025generalreasoneradvancingllmreasoning,huan2025doesmathreasoningimprove}, but still require task-specific verifiers. \textbf{Self-Training. } In parallel, self-training approaches leverage the model’s own outputs as training data. STaR \cite{zelikman2022starbootstrappingreasoningreasoning} bootstraps from reasoning paths that yield correct answers; Self-Refine \cite{madaan2023selfrefineiterativerefinementselffeedback} uses iterative self-critique; and other works exploit self-consistency as a proxy reward for RL \cite{shafayat2025largereasoningmodelsselftrain}, build self-editing/verifying systems \cite{zhang2025darwingodelmachineopenended}, or study the generation–verification gap theoretically \cite{song2025mindgapexaminingselfimprovement,lu2025doesverificationpayoff}. These methods improve reasoning quality by filtering or refining full reasoning paths, but generally treat each correct reasoning in isolation. In contrast, our approach intersects multiple successful reasoning paths to identify \textit{decision pivots}—minimal, verifiable checkpoints that all correct solutions share—and rewrites them into concise short-path reasoning before preference optimization. This distillation step targets reasoning faithfulness and verifiability without relying on handcrafted verifiers or general reasoning metrics \cite{golovneva2023roscoesuitemetricsscoring}.

\vspace{-3mm}
\paragraph{Reasoning Models. }
Chain-of-thought (CoT) prompting \cite{wei2022chainofthought} established that models reason more effectively when they articulate intermediate steps, with later work showing that even a simple prompt (\textit{"Let’s think step by step"}) can elicit this behavior \cite{kojima2022large}. Research since has explored richer reasoning supervision: scratchpad-style intermediates \cite{nye2021show}, self-consistency via majority voting over multiple reasoning paths \cite{wang2023self}, and targeted data curation by aggregating reasoning corpora \cite{Ott_2023}, actively selecting informative queries \cite{diao2024activepromptingchainofthoughtlarge}, or varying difficulty adaptively \cite{fu2022complexity}. Recent trends also include compressing CoTs for efficiency or readability, but these typically focus on token length reduction \cite{munkhbat2025selftrainingelicitsconcisereasoning,hassid2025don,cheng2024compressedchainthoughtefficient} rather than identifying core, verifiable reasoning elements. Our method differs in that it uses multi-path intersection to recover a shared, verifiable reasoning core, rewrites it into a short path form, and uses this distilled form as a supervision signal that bridges the gap between unguided reasoning generation and semi-verifiable reasoning refinement.

\section{Problem Formulation: In search of Short-Path Reasoning}\label{sec:theory}

In this section, we formally introduce the problem setting. Specifically, we consider the general setting where, for each input–output pair $(x,y)$, $x \in \mathcal{X}$ is a question and $y \in \mathcal{Y}$ is its ground-truth answer (e.g., a class label). 
A pre-trained reasoning model $p_{\theta}(r,\hat y \mid x)$ \cite{anthropic2024claude3,yang2025qwen3} jointly generates a chain-of-thought reasoning path $r=(t_{1},t_{2},\dots,t_{l})$ of length $l$ and a prediction $\hat y$ given $x$.
When convenient we use the induced conditionals
$p_{\theta}(y \mid x,r)$ (label probability given a path) and 
$p_{\theta}(r \mid x,y) \propto p_{\theta}(r,y \mid x)$ (reasoning path distribution given the label). We denote the unobserved ground truth reasoning as $r^{\ast}=(t_{1}^{\ast},t_{2}^{\ast},\dots,t_{d}^{\ast})$ with $d \ll l$.
The dataset $\mathcal{D}=\{(x_i,y_i)\}_{i=1}^{N}$ contains $N$ supervised examples. 

Here, the main goal is to produce high-quality reasoning that closely resembles the ideal but unobserved reasoning $r^{\ast}$. More specifically, we aim to obtain reasoning that is (1) correct with respect to the final answer, (2) concise in containing only the information necessary to support that answer, and (3) faithful in the sense of being causally responsible and correct for the model’s prediction. However, several factors make this problem challenging. First, the absence of ground-truth reasoning. While the correct label $y$ is observed, the true reasoning process leading to the label is not. Hence, there is no direct supervision for which intermediate reasoning steps are valid. Second, verifying intermediate reasoning steps is difficult, especially in open-ended reasoning tasks (e.g., general language tasks). Hence, individual reasoning steps are hard to verify, while human verification is costly and lacks scalability. Formally, the objective can be stated as finding a reasoning path $r$ that suffices $\max_{r}\; p_{\theta}\!\bigl(y \mid x, r\bigr),$
subject to $r$ being as concise as possible while preserving correctness. This requires identifying only the key information necessary to support the answer, despite the lack of supervision and step-level verifiability \cite{zhou2025reinforcinggeneralreasoningverifiers}.

\vspace{-3mm}
\paragraph{Positioning: Limitations of Prior Approaches. }
Several existing paradigms attempt to address this challenge but face inherent limitations in this general setting. 
Self-training methods such as STaR~\cite{zelikman2022starbootstrappingreasoningreasoning} bootstrap from model-generated reasoning that yield correct answers, but treat each reasoning path independently and do not explicitly seek minimality or commonality across multiple successful paths. 
Self-consistency~\cite{wang2023self} aggregates predictions from multiple sampled reasoning paths, improving answer accuracy but producing no refined intermediate reasoning for training. 
RLVR~\cite{deepseek2024r1zero,zhang2025darwingodelmachineopenended,zhao2025absolutezeroreinforcedselfplay} uses verifiable rewards to supervise reasoning, but is restricted to domains where intermediate verification is feasible, such as math and code, and does not generalize to open-ended reasoning. 
Other metric-based or LLM-as-judge feedback methods~\cite{golovneva2023roscoesuitemetricsscoring,liu-etal-2023-g,bai2022constitutional} depend on either handcrafted metrics or opaque model judgments, which do not reflect the task. In other words, existing metrics do not reliably capture whether the reasoning truly contains only the information necessary for the decision, which we empirically show in \Cref{sec:experiments}. In the following section, we seek to address the general problem of producing reasoning in the absence of hypothetical ground-truth reasoning and automated verifiers.

\section{Reasoning Optimization via Multi-path Aggregation}\label{sec:method}

\begin{figure}[ht]
    \centering
\includegraphics[width=0.9\textwidth]{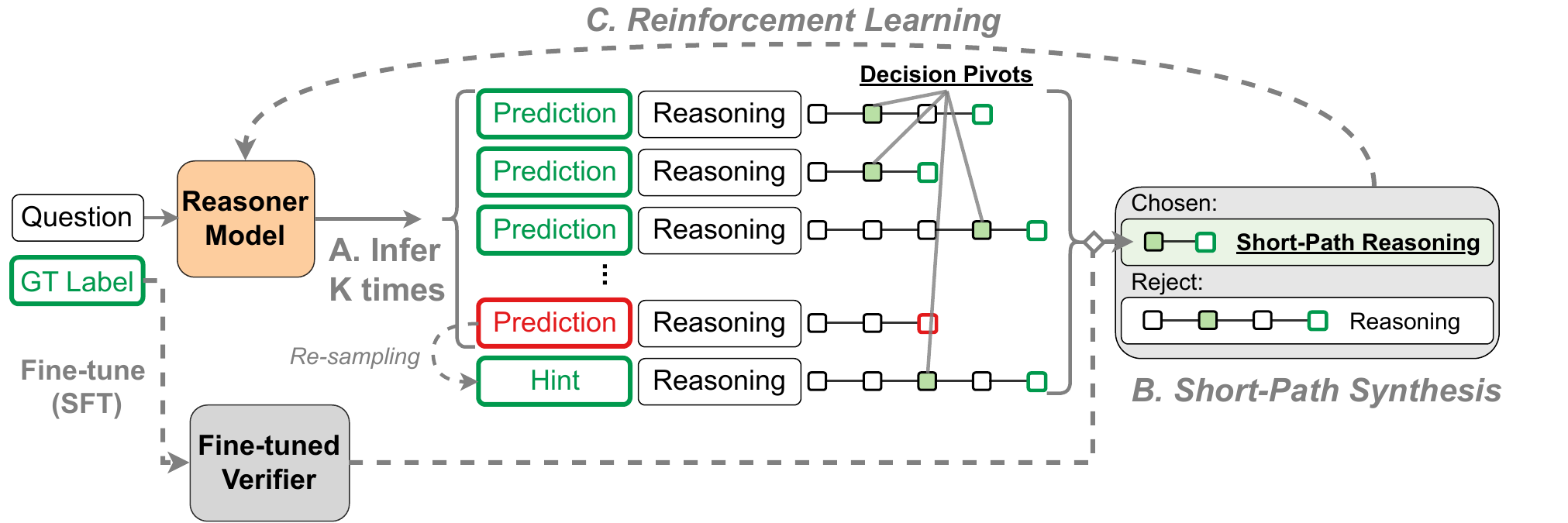}
    \caption{We present a novel self-training framework ROMA that leverages the concept of decision-pivots. Our self-training framework works in 3 stages. Given a question: (A) model produces multiple $\textsc{prediction} + \textsc{reasoning}$ pairs, ensuring we collect $K$ reasoning paths using re-sampling. (B) Then, a fine-tuned verifier synthesizes a short-path reasoning (SPR) that focuses on shared decision-pivots, generating a preference data pair. (C) Use the generated data for Reinforcement Learning (i.e., preference learning with DPO). This process can be repeated as a self-training loop that improves the model's reasoning capabilities.}
    \label{fig:method}
\end{figure}

In this section, we present a self-training framework that bootstraps a concise, human-aligned reasoning path for self-improvement of the model's reasoning capability. The key idea of our method lies in the concept of \textit{decision pivots}, which are key logics / key words that are shared across reasoning paths that reach the same decision (See \Cref{fig:decision-pivots}). Under this assumption, our method bootstraps a pivot-focusing CoT reasoning by distilling multiple reasoning paths into a concise reasoning path that yields the correct prediction (i.e., short-path reasoning -- SPR), then feeds the SPR back into the model to self-improve its reasoning and generalization.

\paragraph{Notation.}
Let $x \in \mathcal{X}$ be an input (e.g., a question) and $y \in \mathcal{Y}$ its ground-truth label.  
We use a pre-trained reasoning model $p_{\theta}(r,\hat y \mid x)$ that, given $x$, jointly produces a reasoning path $r=(t_{1},\dots,t_{l})$ and a prediction $\hat y$ \cite{yang2025qwen3}. 
We write $p_{\theta}(y \mid x,r)$ for the model’s conditional over labels given a reasoning, and $p_{\theta}(r \mid x,y)\propto p_{\theta}(r,y\mid x)$ for the conditional over the reasoning given the label.  
The training corpus is $\mathcal{D}=\{(x_i,y_i)\}_{i=1}^{N}$ of size $N$.  
For each input-label pair $(x_i,y_i)$, define a candidate set of joint samples $\mathcal{S}_i=\{(r_{ik},\hat y_{ik})\}_{k=1}^{K}$ drawn from $p_{\theta}(\cdot\mid x_i)$, and the successful subset $\mathcal{R}_i^{+}=\{\,r_{ik}:\hat y_{ik}=y_i\,\}$. Separate from the reasoner $p_{\theta}$, we adopt a fine-tuned verifier $v_{\psi}:(x,\mathcal{R})\mapsto \hat r$ trained on $\mathcal{D}$ (namely using LoRA adapters \cite{hu2021loralowrankadaptationlarge} on $p_{\theta}$) synthesizes a SPR $\hat r_i=v_{\psi}(x_i,\mathcal{R}_i^{+})$.

\subsection{Stage A: Multi-sample bootstrapping}\label{sec:stage-a}
For each $(x_i,y_i)$, sample $K$ pairs:
\begin{equation}\label{eq:sets}
\mathcal{S}_i=\bigl\{(r_{ik},\hat y_{ik})\sim p_{\theta}(\cdot\mid x_i)\bigr\}_{k=1}^{K}, 
\\
\mathcal{R}_i^{+}=\{\,r_{ik}:\hat y_{ik}=y_i\,\}
\end{equation}
If $\mathcal{R}_i^{+}$ is empty, resample paths from $p_{\theta}(r\mid x_i,y_i)$ to ensure coverage (See Step A in \Cref{fig:method}). This step explores diverse paths while keeping supervision verifiable via the paired prediction; conditioning on $y_i$ functions as a fallback to avoid empty pools. Intuitively, the two sources of reasoning widen the coverage of the model's reasoning. Zero-shot correct reasoning paths reveal the model’s canonical, high-probability pivot sets (i.e., what the model already trusts), while \textit{guided} paths (re-sampled after revealing the hint $y_i$ on the initial incorrect prediction \cite{zelikman2022starbootstrappingreasoningreasoning,amani2025rl}) allow the model to route around its default heuristics and adopt novel paths that rationalize $y_i$.

\subsection{Stage B: Verifier-guided Short-path synthesis}\label{sec:stage-b}

Given a candidate pool of multiple reasoning paths that reach the correct decision $\mathcal{R}_i^{+}$, the verifier $v_{\psi}$ aggregates overlapping content across candidates and produces a refined SPR that focuses on the decision pivots, and preserves the correct decision (See Step B in \Cref{fig:method}):
\begin{equation}\label{eq:verifier}
\hat r_i \;=\; v_{\psi}(x_i,\mathcal{R}_i^{+})
\quad\text{subject to}\quad 
\arg\max_{y} p_{\theta}(y\mid x_i,\hat r_i)=y_i.
\end{equation}
The rewriting aggregates shared decision-pivots while discarding redundant reasoning steps; the correctness check guarantees that the resulting explanation leads to the correct decision. Please note that the verifier $v_{\psi}$ is used only to synthesize $\hat r_i$ and is not our final model.

\vspace{-2mm}
\paragraph{Discussion: Fine-tuned Verifiers.} An important aspect of our method is adopting a fine-tuned verifier $v_{\psi}$ to rewrite SPR, separate from the reasoner $p_{\theta}$ that produces a pool of $K$ candidate paths. This way, the reasoner $p_{\theta}$ remains domain-agnostic, while a lightweight verifier $v_{\psi}$ ($p_{\theta}$ added with LoRA trained on $\mathcal{D}$) acquires domain priors to synthesize SPR during training and is discarded at test time. For clarification, the verifier is trained only on the training split and never applied to test examples. The motivation is the specialization–generalization trade-off: fine-tuning on $\mathcal{D}$ can inject domain knowledge but tends to deteriorate general reasoning ability \cite{chu2025sftmemorizesrlgeneralizes, cho2025forgetforgettingcontinuallearning, wang2024twostagellmfinetuningspecialization, ma2025generalreasoneradvancingllmreasoning}. By isolating this specialization in $v_{\psi}$, we preserve the general capability of $p_{\theta}$ while leveraging the verifier’s domain priors to filter spurious steps, align reasoning paths to decision pivots, and consolidate them into faithful short paths. Here, the cost is minimal as$v_{\psi}$ is obtained via small LoRA adapters on the same backbone. Empirically, verifier-guided rewriting provides stronger learning signals than self-rewriting with $p_{\theta}$ (see Fig.~\ref{fig:verifier} and Tab.~\ref{tab:combined_logiqa_medqa_math} in \Cref{sec:experiments}).

\subsection{Stage C: Pairwise preference optimization}
\label{sec:stage-c}

We align the generator \(p_{\theta}\) to prefer the synthesized short path. For each \((x_i,y_i)\), treat \(\hat r_i\) as chosen and every other successful candidate \(r\in\mathcal{R}_i^{+}\setminus\{\hat r_i\}\) as rejected, forming
\begin{equation}
\mathcal{P}_i=\{(\hat r_i, r): r\in\mathcal{R}_i^{+}\setminus\{\hat r_i\}\}.
\end{equation}
We then optimize a standard DPO objective~\cite{rafailov2023direct} for \(p_{\theta}(r\mid x,y)\) against a frozen reference \(p_{\mathrm{ref}}\) (the exact loss is given in Appendix \ref{appendix:method}).
Intuitively, because \(p_{\theta}(r\mid x,y)=\prod_{t}p_{\theta}(t_t\mid x,y,t_{<t})\), the preference update increases token probabilities precisely at the positions where \(\hat r_i\) and alternatives differ—i.e., the pivot steps—while the implicit KL term (via \(p_{\mathrm{ref}}\)) regularizes drift.

\vspace{-2mm}
\paragraph{Discussion: Pivot-focused supervision.}
Assume each instance $(x,y)$ admits a small pivot set $P(x)$ such that correctness is determined by visiting all pivots.
Let $r \sim p_{\theta}(\cdot \mid x)$ and define $q_t := \Pr_{\theta}[\text{correct token at pivot step }t \in P(x)\mid x]$.
Then the final-error probability satisfies
$\Pr[\hat{y}\neq y] \;\le\; \mathbb{E}_{x}\!\big[\sum_{t\in P(x)}(1-q_t)\big]$,
so reducing the \emph{pivot-miss risk} $\varepsilon_{\mathrm{piv}} := \mathbb{E}_{x}\sum_{t\in P(x)}(1-q_t)$ reduces task error.
Our synthesized SPR $\hat r_i$ differs from other successful paths primarily at pivot-bearing tokens. That is, by the token factorization of $\log p_{\theta}(r\mid x,y)$, the preference update increases $q_t$ exactly at those disagreement positions, concentrating probability mass on pivot-preserving continuations.
In addition, replacing long traces by $\hat r_i$ removes non-pivot tokens and reduces gradient noise away from pivots, further sharpening updates on the few steps that control correctness.
Finally, because preference pairs are built from $p_{\theta}$’s own rollouts, updates act on the prefixes the model actually visits, mitigating distributional shift between training and generation.
For a formal sketch of our analysis, please refer to Appendix~\ref{sec:theory-pivots}.

\section{Experiments}\label{sec:experiments}

\paragraph{Benchmarks.}

We validate our approach across three public benchmarks to demonstrate cross-domain effectiveness. \textsc{LogiQA} \cite{liu2020logiqa} is a general question-answer dataset that assesses logical reasoning with multiple choice questions (MCQs) where the models are evaluated mainly by accuracy. \textsc{MedQA} \cite{jin2021disease} is a medical MCQ benchmark built from professional licensing exams (e.g., USMLE), designed to evaluate clinical knowledge and reasoning based on accuracy. Lastly, \textsc{MATH500} \cite{hendrycks2021measuring} is a mathematical benchmark commonly used to assess a model's reasoning capabilities. On these benchmarks, we compare different types of self-training baselines that utilize their own generated outputs to improve their reasoning capability, which in turn improves a model's downstream task performance \cite{zelikman2022starbootstrappingreasoningreasoning,golovneva2023pathfinder,song2025mindgapexaminingselfimprovement}. In all experiments, we used a held-out set of test data for evaluation. 

\vspace{-3mm}
\paragraph{Architectures and protocol.} We evaluate our method on variants of Qwen-3~\cite{yang2025qwen3}, each of which already demonstrates strong zero-shot reasoning ability. Unless mentioned, the default model is set as the Qwen-3-0.6B model, considering the generation cost/latency of self-training (see \Cref{tab:time}). We further use several larger models (e.g., Claude 3 \cite{anthropic2024claude3}, LLama-3 \cite{grattafiori2024llama3herdmodels}. Every experiment is repeated five times with independent random seeds, and we report the mean performance together with the corresponding standard errors. Details regarding the hyperparameters are presented in the sequel. We assess the effectiveness of our method in two aspects. (1) Downstream Task: Using the generated reasoning, we post-train the model and then assess the trained model on the downstream task (\Cref{tab:combined_logiqa_medqa_math}). (2) Qualitative analysis: Assess the output reasoning using general metrics e.g., ROSCOE\cite{golovneva2023roscoesuitemetricsscoring} (\Cref{tab:roscoe}). Further detail regarding the experimental setting is included in \Cref{appendix:setting}

\subsection{Experimental Results}

\begin{figure}[ht]
    \centering
    \begin{subfigure}[b]{0.49\textwidth}
        \centering
        \includegraphics[width=\textwidth]{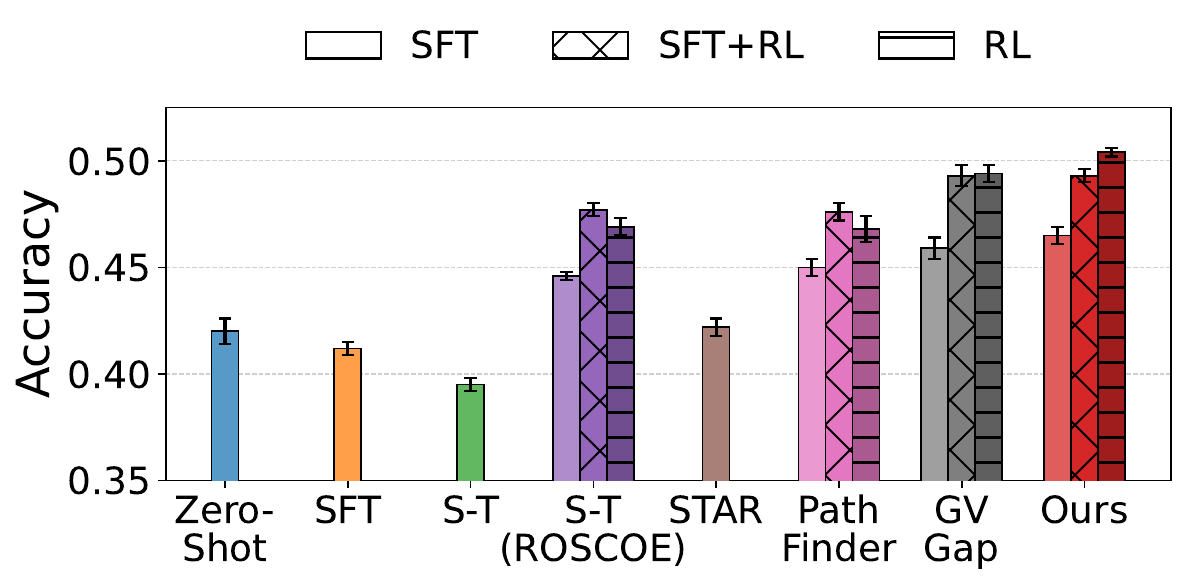}
        \caption{Results in LogiQA}
        \label{fig:logiqa}
    \end{subfigure}
    \hfill
    \begin{subfigure}[b]{0.49\textwidth}
        \centering
        \includegraphics[width=\textwidth]{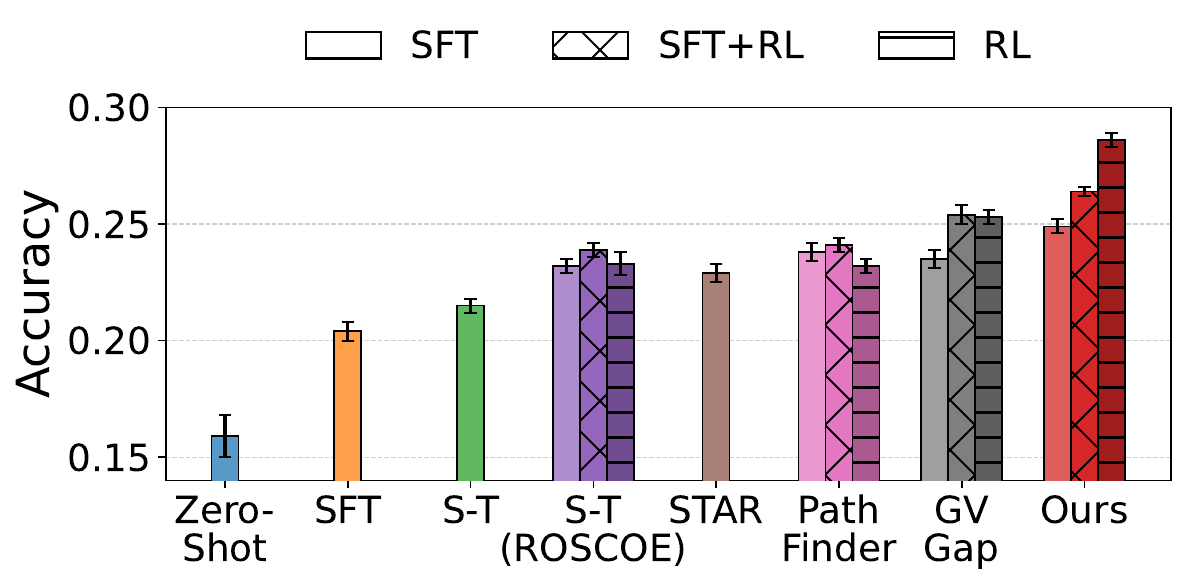}
        \caption{Results in MedQA}
        \label{fig:medqa}
    \end{subfigure}
    \caption{Comparison of self-training results on \textsc{LogiQA} and \textsc{MedQA}. Our proposed method provides large self-improvement gains in domains (e.g., \textsc{MedQA} -- healthcare, \textsc{LogiQA} -- general language) that are generally difficult to verify the generated reasoning, compared to math/coding domains (e.g., \textsc{MATH500} -- math). S-T (ROSCOE) refers to self-training with ROSCOE-filtered reasoning. The error bars indicate the standard error over 5 runs.}
    \label{fig:qa_results}
\end{figure}

We aim to answer the following questions to assess the effectiveness, efficiency, and generality of our proposed method:
\textbf{(1)} Can it generate high-quality reasoning that improves (e.g., after training or zero-shot) generalization?
(\Cref{tab:combined_logiqa_medqa_math})
\textbf{(2)} How do key design components—bootstrapping, rewriting—contribute to its effectiveness? Can the proposed self-training method provide good learning signals continually, improving the model on each loop? (\Cref{tab:ablation_overview})
\textbf{(3)} Do decision pivots truly exist, and are they retrievable? (\Cref{tab:pivot_retrieval})

\begin{table}[t]
\centering
\caption{Accuracy on \text{LogiQA}, \textsc{MedQA}, and \textsc{MATH500} using various types of generated reasoning and self-training methods, across different post-training techniques (e.g., SFT, RL). We report the standard error across 5 runs.}
\label{tab:combined_logiqa_medqa_math}
\adjustbox{max width=\linewidth}{
\begin{tabular}{l l c c c}
\toprule
\textbf{Method} & \textbf{Techniques} & \textbf{LogiQA} & \textbf{MedQA} & \textbf{MATH500} \\
\midrule
Zero-Shot & -- & 0.420$_{\pm{0.006}}$ & 0.159$_{\pm{0.009}}$ & 0.636$_{\pm{0.006}}$ \\
Trained without Reasoning & SFT & 0.412$_{\pm0.003}$ & 0.204$_{\pm0.004}$ & 0.649$_{\pm0.004}$ \\
Self-Trained (Zero-Shot) & SFT & 0.395$_{\pm0.003}$ & 0.215$_{\pm0.003}$ & 0.655$_{\pm0.005}$ \\
\multirow{3}{*}{Self-Trained (ROSCOE [\citenum{golovneva2023roscoesuitemetricsscoring}])} 
 & SFT      & 0.446$_{\pm0.002}$ & 0.232$_{\pm0.003}$ & 0.673$_{\pm0.003}$ \\
 & SFT+RL   & 0.477$_{\pm0.003}$ & 0.239$_{\pm0.003}$ & 0.681$_{\pm0.004}$ \\
 & RL   & 0.469$_{\pm0.004}$ & 0.233$_{\pm0.005}$ & 0.677$_{\pm0.005}$ \\
\midrule
\textsc{star}  [\citenum{zelikman2022starbootstrappingreasoningreasoning}] & SFT & 0.422$_{\pm0.004}$ & 0.229$_{\pm0.004}$ & 0.653$_{\pm0.005}$ \\
\multirow{3}{*}{\textsc{pathfinder} [\citenum{golovneva2023pathfinder}]}  
& SFT      & 0.450$_{\pm0.004}$ & 0.238$_{\pm0.004}$ & 0.668$_{\pm0.005}$ \\
& SFT+RL   & 0.476$_{\pm0.004}$ & 0.241$_{\pm0.003}$ & 0.679$_{\pm0.005}$ \\
& RL & 0.468$_{\pm0.006}$ & 0.232$_{\pm0.003}$ & 0.671$_{\pm0.005}$ \\
\multirow{3}{*}{\textsc{gv-gap} [\citenum{song2025mindgapexaminingselfimprovement}]}  
& SFT      & 0.459$_{\pm0.005}$ & 0.235$_{\pm0.004}$ & 0.670$_{\pm0.003}$ \\
& SFT+RL   & 0.493$_{\pm0.005}$ & $0.254_{\pm0.004}$ & $0.688_{\pm0.004}$ \\
& RL & 0.494$_{\pm0.004}$ & 0.253$_{\pm0.003}$ & 0.686$_{\pm0.004}$ \\
\midrule
\multirow{3}{*}{Ours (w/o verifier)} 
 & SFT      & 0.462$_{\pm0.003}$ & 0.234$_{\pm0.003}$ & 0.669$_{\pm0.004}$ \\
 & SFT+RL   & 0.495$_{\pm0.004}$ & 0.251$_{\pm0.002}$ & 0.684$_{\pm0.003}$ \\
 & RL       & 0.489$_{\pm0.003}$ & 0.250$_{\pm0.003}$ & 0.681$_{\pm0.003}$ \\
\midrule 
\multirow{3}{*}{Ours} 
 & SFT      & 0.465$_{\pm{0.004}}$ & 0.249$_{\pm{0.003}}$ & 0.671$_{\pm{0.004}}$  \\
 & SFT+RL   & 0.493$_{\pm{0.003}}$ & 0.264$_{\pm{0.002}}$ & 0.686$_{\pm{0.002}}$ \\
 & RL       & \textbf{0.504}$_{\pm{0.002}}$ & \textbf{0.286}$_{\pm{0.003}}$ & \textbf{0.692}$_{\pm{0.003}}$ \\
\bottomrule
\end{tabular}
}
\vspace{-3mm}
\end{table}

\paragraph{Main Results. }

Across \textsc{LogiQA}, \textsc{MedQA}, and \textsc{MATH500}, a consistent pattern emerges (Table~\ref{tab:combined_logiqa_medqa_math}): performance improves as supervision becomes more focused on decision pivots rather than full reasoning paths. Training without reasoning supervision or reusing zero-shot reasoning, fails to yield reliable gains and can even degrade performance, indicating that unguided reasoning does not provide a stable learning signal. Simply exposing the model to more reasoning text is therefore insufficient; without guidance, the model cannot distinguish decision-critical steps from redundant or spurious ones. Filtering self-generated reasoning using general-purpose quality metrics (e.g., ROSCOE) partially alleviates this issue by removing low-quality paths, but its effect plateaus. Such metrics capture surface-level coherence or fluency, yet remain agnostic to task-specific decision pivots, limiting their ability to drive further improvements. 

In contrast, compressing multiple successful reasoning paths into a pivot-focused short-path reasoning (SPR) concentrates supervision on the minimal information shared across correct solutions. When paired with preference learning, this pivot-centric supervision consistently outperforms metric-filtered and self-training baselines across all benchmarks. The strongest gains arise when SPR synthesis is guided by a domain-tuned verifier (\cref{sec:stage-b}) , rather than the reasoner itself. The verifier injects task-specific priors, prunes spurious steps, and sharpens the preference signal during consolidation. This effect scales with domain knowledge requirements: it is most pronounced on \textsc{MedQA}, clear on \textsc{LogiQA}, and smaller yet consistent on \textsc{MATH500}, where intermediate correctness is already externally verifiable.

These empirical trends directly support our theoretical analysis in Appendix~\ref{sec:theory-pivots}, which formalizes decision pivots as sparse, verifiable determinants of correctness. By concentrating supervision on pivot-satisfying paths and amplifying their influence through verifier-guided consolidation, our method increases the alignment between training signals and task utility, explaining both the robustness and the domain-dependent scaling of the observed gains.

\paragraph{Effect of Fine-tuned Verifiers.}

\begin{figure}[ht]
    \centering
\includegraphics[width=0.99\textwidth]{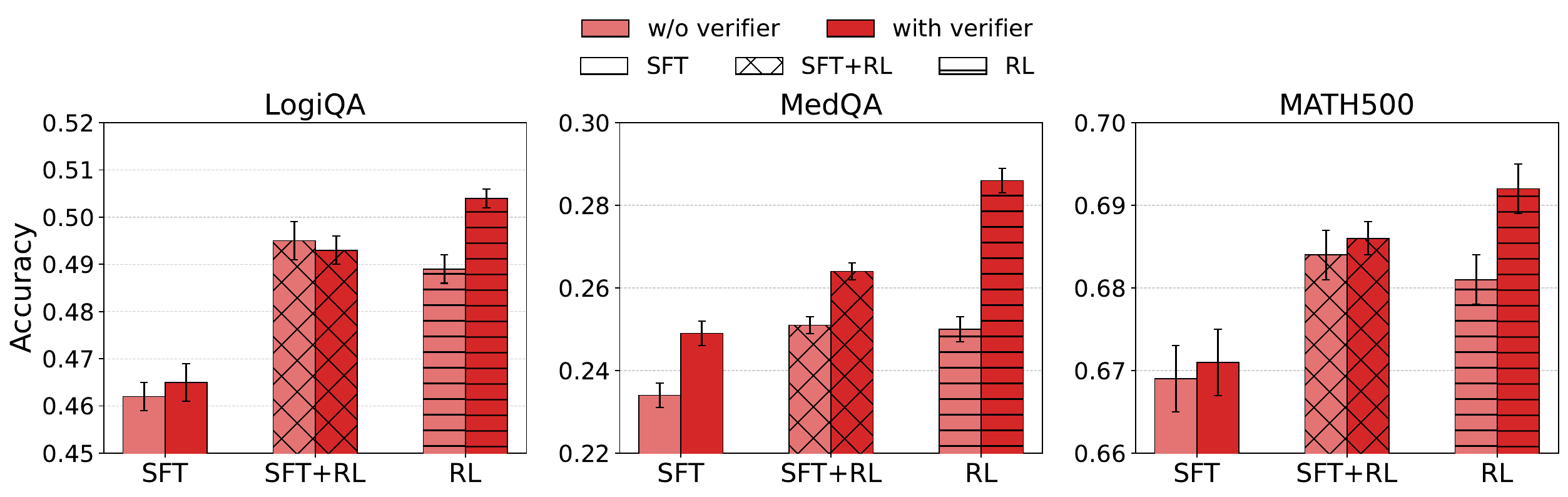}
    \caption{Effect of the fine-tuned verifier on downstream accuracy across post-training regimes. We compare our method with and without a verifier on LogiQA, MedQA, and MATH500. The verifier yields consistent gains and is most impactful in the expert MedQA domain, supporting our claim that a domain-tuned verifier synthesizes higher-quality, pivot-focused SPR. Improvements are smaller on MATH500 (where an external verification signal already exists) and modest on LogiQA, aligning with our hypothesis.}
    \label{fig:verifier}
\end{figure}

Incorporating a fine-tuned verifier consistently improves downstream accuracy by synthesizing higher-quality, pivot-focused SPR (Table~\ref{tab:combined_logiqa_medqa_math}; Figure~\ref{fig:verifier}). Across SFT, SFT+RL, and RL regimes, the verifier refines supervision by filtering spurious steps, aligning paths to shared decision pivots, and injecting domain-specific priors during consolidation, resulting in a sharper and more informative learning signal. The magnitude of this effect follows task characteristics. Gains are largest on the knowledge-intensive \textsc{MedQA} benchmark, where domain priors are crucial for identifying valid pivots. Improvements are smaller but consistent on \textsc{MATH500}, where intermediate correctness is already externally verifiable, and more modest on \textsc{LogiQA}, which involves weaker domain constraints and higher reasoning ambiguity~\cite{wu2025knowledgereasoningcloselook}. In practice, we implement the verifier with LoRA adapters \cite{hu2021loralowrankadaptationlarge} for a favorable accuracy–compute trade-off; when domains shift, we retune the verifier to refresh its priors (\Cref{tab:combined_logiqa_medqa_math}). Ultimately, these results align with our analysis in Appendix~\ref{sec:theory-pivots}, which predicts larger gains when correctness depends on a sparse, domain-sensitive set of decision pivots that are easier to verify than to generate.

\paragraph{Are General Reasoning Metrics sufficient for preference data selection?}

\begin{figure}[t]
\centering
\begin{minipage}[t]{0.49\textwidth}
\centering
\vspace{0pt}
\captionof{table}{ROSCOE scores and the downstream LogiQA accuracy across model sizes.}
\label{tab:roscoe}
\fontsize{9}{10}\selectfont
\adjustbox{max width=0.95\linewidth}{
\centering
\begin{tabular}{lrrrrr}
\toprule
Metric & 0.6B & 1.7B & 4B & 8B & 14B \\
\midrule
Faithfulness & 0.902 & 0.872 & 0.885 & 0.892 & 0.888 \\
Informativeness (Step) & 0.893 & 0.875 & 0.886 & 0.889 & 0.883 \\
Informativeness (Chain) & 0.934 & 0.938 & 0.944 & 0.928 & 0.924 \\
Faithfulness (WW) & 0.948 & 0.941 & 0.944 & 0.940 & 0.939 \\
Repetition (Word) & 0.023 & 0.012 & 0.009 & 0.018 & 0.021 \\
Repetition (Step) & 0.020 & 0.012 & 0.014 & 0.016 & 0.019 \\
Discourse Rep. & 0.007 & 0.002 & 0.001 & 0.002 & 0.004 \\
Coherence & 0.004 & 0.001 & 0.002 & 0.001 & 0.052 \\
Perplexity (Step) & 0.002 & 0.005 & 0.006 & 0.004 & 0.005 \\
Perplexity (Chain) & 0.068 & 0.128 & 0.116 & 0.076 & 0.084 \\
Grammar (Step) & 0.901 & 0.891 & 0.889 & 0.894 & 0.890 \\
Grammar (Step Max) & 0.557 & 0.355 & 0.371 & 0.339 & 0.442 \\
\midrule
Accuracy & 0.552 & 0.486 & 0.629 & 0.664 & 0.672 \\
\bottomrule
\end{tabular}
}
\end{minipage}
\hfill
\begin{minipage}[t]{0.49\textwidth}
\centering
\vspace{0pt}
\includegraphics[width=\linewidth]{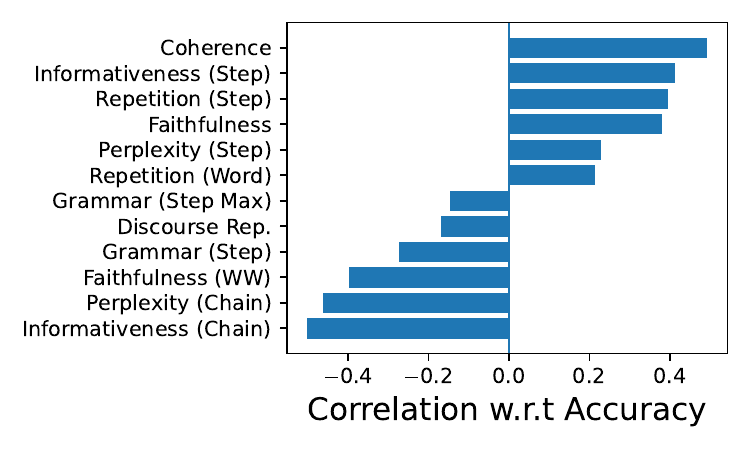}
\captionof{figure}{Pearson Correlation between ROSCOE scores and downstream LogiQA accuracy.}
\label{fig:roscoe_corr}
\end{minipage}
\end{figure}

\Cref{tab:roscoe} and \Cref{fig:roscoe_corr} show that general-purpose reasoning metrics such as ROSCOE are poorly aligned with downstream task accuracy, making them unreliable signals for preference data selection. We observe three consistent patterns. \textbf{(1)} As model size increases, most ROSCOE facets remain nearly flat or exhibit idiosyncratic behavior, while downstream accuracy improves substantially (Table~\ref{tab:roscoe}). \textbf{(2)} Cross-scale correlations make this misalignment explicit: several ROSCOE scores are weakly correlated or even negatively correlated with accuracy (e.g., \textsc{Informativeness (Chain)}, \textsc{Faithfulness (WW)}), and the strongest positive correlations are only modest (Figure~\ref{fig:roscoe_corr}). \textbf{(3)} This “flat-high” profile indicates ceiling effects that obscure the variations most predictive of correctness.

This behavior is expected by design. ROSCOE emphasizes fluency and  consistency of reasoning chains, not whether they are sufficiently logical or factual to reach the correct answer. As a result, selecting preference data using ROSCOE removes clearly flawed paths at a general level but fails to prioritize the steps that truly rationalize the decision. In contrast, aggregating multiple correct trajectories into SPR explicitly targets shared decision pivots and when combined with a lightweight verifier, it yields larger and more reliable gains than metric-based selection.

\paragraph{Ablation Study.}

\begin{table}[t]
\centering
\caption{Ablation study of our method. (Left) Varying the number of candidate reasoning paths $K$; (Right) varying the number of self-training loops.}
\label{tab:ablation_overview}
\setlength{\tabcolsep}{6pt}

\begin{subtable}[t]{0.48\textwidth}
\centering
\caption{Effect of $K$ (candidate pool size).}
\label{tab:ablation_K}
\adjustbox{max width=\linewidth}{\begin{tabular}{lcccc}
\toprule
Method & $K$ & LogiQA & MedQA & MATH500 \\
\midrule
Ours & 1  & 0.437 & 0.252 & 0.649 \\
Ours & 3  & 0.499 & 0.273 & 0.680 \\
Ours & 5  & 0.504 & 0.286 & 0.692 \\
Ours & 7  & 0.509 & 0.292 & 0.695 \\
Ours & 10 & \textbf{0.511} & \textbf{0.296} & \textbf{0.699} \\
\bottomrule
\end{tabular}}
\end{subtable}
\hfill
\begin{subtable}[t]{0.48\textwidth}
\centering
\caption{Effect of the number of self-training loops.}
\label{tab:ablation_loops}
\adjustbox{max width=\linewidth}{\begin{tabular}{lcccc}
\toprule
Method & Loops & LogiQA & MedQA & MATH500 \\
\midrule
Ours & 1 & 0.504 & 0.286 & 0.692 \\
Ours & 2 & 0.522 & 0.290 & \textbf{0.699} \\
Ours & 3 & 0.525 & 0.294 & 0.683 \\
Ours & 4 & \textbf{0.526} & 0.297 & 0.694 \\
Ours & 5 & 0.522 & \textbf{0.297} & 0.693 \\
\bottomrule
\end{tabular}}
\end{subtable}

\vspace{-2mm}
\end{table}

We analyze two key aspects: (1) the number of candidate reasoning paths $K$ used to synthesize SPR, and (2) the number of self-training loops. \textbf{Candidate pool size ($K$).}
As shown in Table~\ref{tab:ablation_K}, increasing $K$ consistently improves performance across all benchmarks, with the largest gains occurring when moving from $K{=}1$ to $K{=}3$. This reflects improved pivot coverage and reduced sensitivity to peculiar single-paths. Performance continues to improve with larger $K$ but exhibits diminishing returns beyond $K{=}5$, indicating that the verifier can effectively consolidate a small, diverse set of paths into an equally informative short path. In practice, $K{=}7$ recovers approximately $95$--$100\%$ of the performance achieved at $K{=}10$ while substantially reducing sampling and verification cost; we therefore adopt $K{=}7$ as the default. \textbf{Self-training loops.}
Table~\ref{tab:ablation_loops} shows steady gains from one to two self-training loops as the model internalizes the synthesized short-path signals. A third loop yields smaller but generally positive improvements, while additional loops lead to saturation and occasional regression on math-heavy tasks. We attribute this behavior to pivot-set saturation and mild distributional drift toward synthesized paths, suggesting that a small number of loops is sufficient to capture most of the benefit. We therefore use three loops in the main results, with early stopping on validation accuracy to avoid overfitting to the generated distribution. To sum up, the takeaways are as follows:
(i) Larger candidate pools primarily help by increasing pivot diversity rather than sheer quantity; once the verifier sees enough distinct routes, additional paths add little value. (ii) Iterative self-training is beneficial but exhibits diminishing returns; modest loop counts (2--3) strike a good balance between quality and compute.

\begin{table}[t]
\centering
\caption{Pivot retrieval rate (\%) as a function of model size and number of zero-shot candidate reasoning samples $K$ per prompt. A retrieval is counted when the generated candiate reasoning visits (i.e., mentions) the human-annotated decision pivot set.}
\label{tab:pivot_retrieval}
\adjustbox{max width=\linewidth}{\begin{tabular}{lccc}
\toprule
\textbf{Model} & \textbf{$K{=}10$} & \textbf{$K{=}100$} & \textbf{$K{=}500$} \\
\midrule
Qwen-3 0.6B & 87\% & 89\% & 93\% \\
Qwen-3 1.7B & 85\% & 92\% & 95\% \\
Qwen-3 4B   & 90\% & 95\% & 97\% \\
Qwen-3 8B   & 94\% & 95\% & 97\% \\
Qwen-3 14B  & 96\% & 96\% & 97\% \\
Llama-3-70B & 94\% & 95\% & 96\% \\
Claude-3.7-Sonnet & 95\% & 96\% & 95\%\\
\bottomrule
\end{tabular}}
\vspace{-3mm}
\end{table}

\paragraph{Qualitative Analysis on Decision Pivots.}

Our method assumes that correct solutions repeatedly traverse a small set of decision pivots and that a reasoning path must visit these pivots to reach the correct decision. To probe this assumption, we construct a human-labeled set of 500 Q\&A items with pivot annotations (keywords or key logical steps justifying the answer). For each item, we draw $K$ zero-shot samples and compute a pivot-retrieval rate: the fraction of items for which at least one of the $K$ sampled paths explicitly \emph{visits} the human-labeled pivot.\footnote{
A reasoning path is said to visit a pivot if it explicitly mentions a lemmatized text span corresponding to that pivot in the annotated lexicon. We apply normalization and disambiguation rules (e.g., synonym merging and context checks) as detailed in Appendix~\ref{sec:theory-pivots}.
} As shown in \Cref{tab:pivot_retrieval}, retrieval generally \emph{increases and then saturates} with $K$, with the largest gains from $K{=}10$ to $K{=}100$ and smaller increments beyond. Larger models tend to start closer to saturation even at $K{=}10$ (e.g., $8$–$14$B $\sim$94--96\%), while smaller models benefit more from additional resampling. Notably, even a 0.6B model recovers human pivots with high probability under modest resampling, supporting the view that correct traces repeatedly visit a sparse, shared set of pivots. Appendix~\ref{sec:theory-pivots} connects these observations to our pivot-risk analysis and includes qualitative examples.

\paragraph{Effect on reasoning length without length incentives.}

\begin{table}[t]
\centering
\caption{Effect of generated reasoning on the trained model's average output reasoning length across 1.5k samples.}
\label{tab:length}
\adjustbox{max width=\linewidth}{\begin{tabular}{l r}
\toprule
\textbf{Trained with.} & \textbf{Average \#. of Tokens} \\
\midrule
Zero-Shot Reasoning & 1184.46 \\
Metric-filtered Reasoning & 1038.82 \\
\textsc{star} [\citenum{zelikman2022starbootstrappingreasoningreasoning}] & 1060.95 \\
\textsc{pathfinder} [\citenum{golovneva2023pathfinder}] & 761.53 \\
Short-Path Reasoning (Ours) & \textbf{750.40} \\
\bottomrule
\end{tabular}}
\end{table}


We study how training data type affects the model’s reasoning length under identical decoding, with no explicit incentives to be brief. Table~\ref{tab:length} shows that pivot-focused supervision yields markedly more compact generations. Relative to zero-shot (1184.46 tokens on average), Max-ROSCOE reduces length to 1038.82 ($-145.64$, $-12.3\%$), while our short-path reasoning (SPR) compresses further to 750.40 ($-434.06$, $-36.7\%$). SPR is also $-288.42$ tokens ($-27.8\%$) shorter than Max-ROSCOE without any length filtering or penalties, supporting the claim that focusing on decision pivots naturally shortens the path. These results align with recent observations that self-training elicits shorter reasoning outputs~\cite{munkhbat2025selftrainingelicitsconcisereasoning}. In our case, the pivot-focused SPR teacher offers a simple mechanism for this effect by pruning non-pivot tokens and lowering the teacher’s entropy, so the model learns to prefer shorter paths even without any length incentives.

\paragraph{Decision Pivots for Reasoning Evaluation and Selection. }

\begin{table}[t]
\centering
\caption{Comparison of different preference data selection methods on \textsc{LogiQA}, \textsc{MedQA}, and \textsc{MATH500}. We report the downstream task accuracy after RL (DPO) training.}
\label{tab:pivot_metric}
\adjustbox{max width=\linewidth}{\begin{tabular}{l c c c}
\toprule
\textbf{Preference Data} & \textbf{LogiQA} & \textbf{MedQA} & \textbf{MATH500} \\
\midrule
N/A (Zero-Shot) & 0.420$_{\pm{0.006}}$ & 0.159$_{\pm{0.009}}$ & 0.636$_{\pm{0.006}}$ \\
ROSCOE [\citenum{golovneva2023roscoesuitemetricsscoring}] & 0.469$_{\pm0.004}$ & 0.233$_{\pm0.005}$ & 0.677$_{\pm0.005}$ \\
\midrule
PAS (Ours) & 0.466$_{\pm0.003}$ & 0.236$_{\pm0.004}$ & 0.680$_{\pm0.003}$ \\
PAS + ROSCOE & \textbf{0.470}$_{\pm0.004}$ & \textbf{0.238}$_{\pm0.003}$ & \textbf{0.682}$_{\pm0.003}$ \\

\bottomrule
\end{tabular}

}
\vspace{-3mm}
\end{table}

We introduce a pivot-based verifiability score for reasoning assessment. Given a pool of $K$ candidates $\{r_k\}$ and the short-path $\hat r$ (our best available reasoning), let $P(r)$ denote pivots extracted and validated by the verifier $v_{\psi}$ from path $r$, and let $P^\star = P(\hat r)$. Importantly, $\hat r$ serves only as a reference consolidation of shared decision pivots, not as a correctness oracle. We score each candidate by the F\(_1\) overlap with the short-path pivots,
\(\frac{2\,|P(r)\cap P^\star|}{|P(r)|+|P^\star|+\varepsilon}\) (with small $\varepsilon>0$ for stability). The intuition is that $\hat r$ consolidates the necessary decision pivots, so a verifiable candidate should cover most of $P^\star$ while avoiding spurious pivots. In practice, we use $v_{\psi}$ (LLM-as-judge \cite{bai2022constitutional}) for the scoring. To avoid circularity and judge bias, we use a frozen, cross-family verifier for pivot extraction and require structured outputs with evidence spans. Hence, extractions with low confidence or missing evidence are discarded. For preference-data selection (\Cref{tab:pivot_metric}), the pivot score alone matches ROSCOE on \textsc{LogiQA} (0.466 vs.\ 0.469) and yields modest gains on \textsc{MedQA} (0.236 vs.\ 0.233) and \textsc{MATH500} (0.680 vs.\ 0.677). Crucially, combining the pivot score with ROSCOE delivers the best results across all three benchmarks (0.470 / 0.238 / 0.682), indicating that verifiability via decision pivots is complementary to general CoT quality metrics and leads to more effective data curation. While our experiments employ DPO, the signal is algorithm-agnostic and can extend to other RLHF variants (e.g., GRPO, PPO) for both offline selection and on-policy filtering.

\vspace{-4mm}
\section{Concluding Remarks}\label{sec:conclusion}
\vspace{-3mm}
Decision pivots provide a principled lens for improving and refining reasoning models at scale. By mining these pivots and compressing explanations into short-path reasoning, our self-training framework delivers concurrent gains in task accuracy and reasoning quality, all without human-annotated reasoning. Crucially, we decouple \textit{reasoning} and \textit{rewriting}: the reasoner supplies diverse reasoning paths while a lightweight, domain-tuned verifier rewrites them. Although such fine-tuned verifiers tend to sacrifice broad, general-purpose reasoning ability, they rewrite more precisely within their domain; using the reasoner for coverage and the verifier for consolidation makes the resulting traces more verifiable, pivot-focused, and ultimately more effective supervision. The improvements on multiple benchmarks (e.g., LogiQA, MedQA, MATH500) suggest that the concept decision pivots are broadly applicable and open a path toward human-level reasoning.

\bibliography{main}
\bibliographystyle{plainnat}

\appendix
\newpage
\section{Appendix}

\subsection{Discussion on the Proposed Method}\label{appendix:method}

\begin{tcolorbox}[
  colback=gray!5, colframe=black!30,
  boxrule=0.5pt, arc=6pt,
  left=8pt, right=8pt, top=6pt, bottom=6pt,
  width=\linewidth
]
\captionsetup{type=listing}
\captionof{listing}{Prompt for consolidating $K$ reasoning paths into a single refined Short-Path Reasoning (SPR).}
\label{lst:refine_prompt}

\begin{Verbatim}[fontsize=\small, breaklines=true, breakanywhere=true]
You are an expert analyst. You will be given several correct reasoning paths. Rewrite a refined reasoning that focuses on shared key information/keywords.

I have the following reasoning paths:
{chr(10).join([f"Reasoning {i+1}: {r}" for i, r in enumerate(k_reasonings)])}

Begin by providing a list of shared decision pivots. Only include the decision pivots when they are visited by the majority of the provided reasoning paths in the candidate pool.

Then, aggregate the multiple reasoning paths and provide a single refined reasoning that:

1) focuses on the decision pivots (keywords/key information) that are shared across the candidate paths
2) Avoids repetition
\end{Verbatim}
\end{tcolorbox}

\paragraph{Intuition.}
Reasoning paths that reach the same correct decision tend to converge on a small set of decision pivots—minimal, verifiable facts or logical checkpoints that are causally responsible for the answer—while differing in style and superfluous steps. Our method exploits this structure: (i) surface diverse but successful paths, (ii) intersect them to recover the shared pivots, and (iii) rewrite a compact Short-Path Reasoning (SPR) that keeps only pivot-bearing content. Preference learning then aligns the model toward these concise, faithful explanations.

\paragraph{Backbone and prompting.}
Unless noted, the generator $p_{\theta}$ is a [backbone name; e.g., Qwen-3~\cite{yang2025qwen3}] with a context window of $[X]$ tokens.
We use a two-channel prompting scheme (\textit{thinking} for internal draft, \textit{answer} for external output) and strip any internal paths at training time except when explicitly collected in Stage~A.

\paragraph{Stage A: Multi-sample bootstrapping (diversity with guardrails).}
For each $(x_i, y_i)$, we obtain a candidate pool of size $K$ by interleaving zero-shot sampling and a guided fallback:
\begin{enumerate}\itemsep2pt
\item Zero-shot pass. Prompt the model to predict the label. It returns thinking (reasoning for its prediction) plus result (predicted label). If the predicted label matches $y_i$, we save the thinking path into the pool $R_i^+$; these paths reflect the model’s canonical high-probability pivots.
\item Guided fallback on errors. If the prediction is incorrect, we provide the ground-truth label $y_i$ and ask the model to justify it step-by-step (without revealing that $y_i$ was given as a hint). The model again emits thinking plus result (an external, stepwise explanation for $y_i$). To avoid hint-contamination of internals, we discard the thinking and save only the external explanation (result) into $R_i^+$. These guided paths widen pivot coverage by routing around the model’s default heuristics.
\end{enumerate}
We repeat until $|R_i^+|=K$ (resampling as needed). This produces a pool of successful paths that agree on the label while differing in which pivots they articulate and how they organize them.

\paragraph{Stage B: Short-path synthesis with a lightweight verifier (pivot recovery).}
Given $R_i^+$, a small verifier $v_\psi$ (a LoRA-tuned adapter on the same backbone) aggregates overlapping content and rewrites a single concise reasoning $\hat r_i$ that (a) enumerates the shared decision pivots visited by the majority of candidates and (b) produces a compact chain that foregrounds only those pivots. We use the prompt in \Cref{lst:refine_prompt}, summarized as:
\begin{itemize}\itemsep2pt
\item Step 1 (pivot mining). List the shared decision pivots and include only those visited by the majority of paths. This majority rule suppresses idiosyncratic steps and raises the signal-to-noise ratio of the pivot set.
\item Step 2 (pivot-centric rewrite). Aggregate the paths and provide a single refined reasoning that focuses on shared pivots and avoids repetition. This enforces minimality and fluency while preserving causal sufficiency.
\end{itemize}
Finally, we verify that $\arg\max_y p_\theta(y\mid x_i,\hat r_i)=y_i$ before accepting $\hat r_i$ (a simple correctness check). The verifier is used only at training time for synthesis; it is discarded at test time to keep inference cost unchanged. See also \Cref{lst:refine_prompt} for the exact template.

\paragraph{Stage C: Preference optimization (DPO over pairs).}

\begin{equation}
\label{eq:dpo}
\begin{split}
\mathcal{L}_{\mathrm{DPO}}(\theta)
&=
- \sum_{i=1}^{N}
  \sum_{(r^{+}, r^{-}) \in \mathcal{P}_i}
\log \sigma\Big(
\beta \Big[
\\
&\quad
\underbrace{\log p_{\theta}(r^{+}\!\mid x_i, y_i)
- \log p_{\theta}(r^{-}\!\mid x_i, y_i)}_{\text{policy preference}}
\\
&\quad
-\;
\underbrace{\log p_{\mathrm{ref}}(r^{+}\!\mid x_i, y_i)
- \log p_{\mathrm{ref}}(r^{-}\!\mid x_i, y_i)}_{\text{implicit KL correction}}
\Big]
\Big)
\end{split}
\end{equation}

We treat the synthesized $\hat r_i$ as chosen and each remaining $r\in R_i^+\setminus\{\hat r_i\}$ as rejected, forming pairs $(\hat r_i, r)$. Direct Preference Optimization (DPO) (see \Cref{eq:dpo}) then increases the relative likelihood of $\hat r_i$ under the policy while regularizing toward a frozen reference. Intuitively, DPO turns the pivot-dense rewrite into a stronger learning signal than any single raw path.

\paragraph{How each component contributes.}
(i) Bootstrapping supplies diverse, label-consistent evidence, mixing the model’s canonical routes (zero-shot) with counter-heuristic routes (guided) to stabilize pivot extraction. (ii) Verifier-guided synthesis converts many paths into one faithful, minimal SPR by explicitly mining majority pivots and pruning spurious steps; this raises pivot density and tightens decision margins. (iii) Preference learning aligns the generator toward SPR: rewarding causal sufficiency and compactness rather than length or stylistic goodness—and does so without hand-crafted external metrics or human rationales.

\vspace{2mm}
\noindent Prompt for Step B (exact). See \Cref{lst:refine_prompt} for the full template used to (1) list shared pivots and (2) produce the refined SPR. This two-stage instruction was crucial in practice: eliciting pivots before rewriting makes the subsequent consolidation measurably more stable than asking for a single rewrite in one shot.

\vspace{1mm}
\noindent Remark on modes. Thinking mode~\cite{yang2025qwen3} is leveraged only to collect richer raw material in Stage A (and we carefully choose which channel to retain); the final trained model can be deployed with or without thinking mode at inference, since the verifier is not required at test time.


\subsection{Theoretical Analysis on Decision Pivots}
\label{sec:theory-pivots}
In this section, we provide a theoretical analysis of why pivot-focused short-path reasoning (\textsc{spr}) improves self-training. The analysis combines two established perspectives: (i) the generation--verification gap (GV-gap)~\cite{song2025mindgapexaminingselfimprovement} framework, which characterizes the effect of reweighting self-generated samples using a verifier, and (ii) a distillation viewpoint, where supervision concentrated on decision-critical tokens yields lower-variance and more sample-efficient updates~\cite{gu2025minillmknowledgedistillationlarge}.

\paragraph{Setup and notation.}
For each input $x$, a reasoner $p_\theta(r,\hat y\mid x)$ generates a reasoning path $r=(t_1,\dots,t_\ell)$ and an answer $\hat y$.  
Let $s(x,r)\in[0,1]$ denote task success or utility~\cite{song2025mindgapexaminingselfimprovement} (e.g., correctness), and define the expected success
\begin{equation}
J(\theta) := \mathbb{E}_{x}\,\mathbb{E}_{r\sim p_\theta(\cdot\mid x)}\!\big[s(x,r)\big].
\end{equation}
A verifier $v_\psi(x,r)\in[0,1]$ induces a weighting function $w(v_\psi(x,r))\ge 0$ (i.e., nonnegative),
which defines the reweighted distribution
\begin{equation}
p^{(w)}_\theta(r\mid x) \;=\;
\frac{p_\theta(r\mid x)\,w(v_\psi(x,r))}
{\mathbb{E}_{\tilde r\sim p_\theta(\cdot\mid x)}[w(v_\psi(x,\tilde r))]}.
\end{equation}
Here, $v_\psi(x,r)$ denotes a scalar verifier score measuring the quality or
correctness of a reasoning path, and $w(\cdot)$ maps this score to a nonnegative
training weight. While $p^{(w)}_\theta(\cdot\mid x)$ is introduced for analysis, in our method the verifier uses its scores to select and consolidate multiple candidate paths into
a single pivot-preserving short-path reasoning (\textsc{spr}), which then serves
as the distillation target. For clarification, however, while pivots are conceptual decision-critical units, we operationalize their detection via normalized surface forms for evaluation.

\paragraph{Verifier reweighting via the GV-gap.}
Under this setting, the effect of verifier-based reweighting (i.e., assigning more probability mass to certain paths and less to others; preference weighting) on expected task
performance (see \Cref{fig:verifier} and \Cref{tab:combined_logiqa_medqa_math})
can be expressed by the following identity:
\begin{equation}
J^{(w)}(\theta)-J(\theta)
= \mathbb{E}_x\!\left[
\frac{\operatorname{Cov}_{r\sim p_\theta(\cdot\mid x)}\big(s(x,r),\,w(v_\psi(x,r))\big)}
{\mathbb{E}_{r\sim p_\theta(\cdot\mid x)}[w(v_\psi(x,r))]}
\right].
\label{eq:gvgap}
\end{equation}
Thus, whenever the verifier score $w\!\circ v_\psi$ is positively correlated with task success under the current model’s generated reasoning outputs, reweighting improves expected performance. This captures the standard GV-gap intuition that verification is beneficial when it correlates with correctness~\cite{song2025mindgapexaminingselfimprovement,saadfalcon2025shrinkinggenerationverificationgapweak}.

\paragraph{Pivot bottleneck assumption.}
We assume each instance $x$ admits a small set of decision pivots $P(x)$ (not necessarily unique) such that missing one or more pivots constitutes a dominant failure mode. Define the indicator
\begin{equation}
H_x(r) := \mathbf{1}\{\text{$r$ visits all required pivots in }P(x)\}.
\end{equation}
We decompose task success as
\begin{equation}
s(x,r) = H_x(r)\cdot \tilde s(x,r), \qquad \tilde s(x,r)\in[0,1].
\label{eq:pivot-decomp}
\end{equation}
This modeling assumption captures tasks where correctness is bottlenecked by a small number of necessary reasoning events. Our empirical pivot-retrieval results (\Cref{tab:pivot_retrieval}) support that such pivots are repeatedly expressed across successful reasoning paths.

\paragraph{A sufficient condition for positive GV-gap (sketch).}
Fix $x$ and let $q(x):=\Pr_{r\sim p_\theta(\cdot\mid x)}[H_x(r)=1]$. Suppose that
\begin{equation}
\Delta_s(x):=\mathbb{E}[\tilde s\mid H_x{=}1]-\mathbb{E}[\tilde s\mid H_x{=}0] \ge \delta>0,
\qquad
\Delta_w(x):=\mathbb{E}[w(v_\psi)\mid H_x{=}1]-\mathbb{E}[w(v_\psi)\mid H_x{=}0] \ge \eta>0.
\end{equation}
Then, by a two-group covariance decomposition over $\{H_x{=}1,H_x{=}0\}$,
\begin{equation}
\operatorname{Cov}_{r}\big(s(x,r),w(v_\psi(x,r))\big)
\;\ge\; \delta\,\eta\,q(x)\big(1-q(x)\big).
\end{equation}
Combining this bound with~\eqref{eq:gvgap} yields $J^{(w)}(\theta)-J(\theta)>0$. We emphasize that this provides a sufficient condition explaining when verifier-guided reweighting is expected to help; it is not a universal guarantee.

\paragraph{Why pivot-focused short paths sharpen distillation.}
Our \textsc{spr} teacher distribution $p_{\textsc{spr}}(r\mid x)$ concentrates probability mass on short, pivot-preserving paths synthesized from correct candidates. Distillation by minimizing
\begin{equation}
\mathbb{E}_x\,\mathrm{KL}\!\left(p_{\textsc{spr}}(\cdot\mid x)\,\|\,p_\theta(\cdot\mid x)\right)
\end{equation}
reduces to cross-entropy under $p_{\textsc{spr}}$. Ultimatery, by pruning non-pivot detours, \textsc{spr} increases the density of pivot-bearing tokens in the supervision signal, concentrating gradient updates on decision-critical positions where candidate paths disagree. In contrast, distilling from longer, candidate reasoning chains allocates substantial gradient mass to stylistic or redundant tokens that are weakly coupled to correctness. This provides an optimization-level explanation for the empirical stability and efficiency gains of \textsc{spr} shown in \Cref{sec:experiments}. Yet, we do not claim all reasoning tasks admit a strict pivot bottleneck. Rather, our above analysis explains why the approach is effective when such structure exists, as empirically shown in \Cref{sec:experiments}.

\subsection{Experimental Results and Analysis}\label{appendix:analysis}

\paragraph{Does reasoning affect downstream task performance?}

\begin{table}[h!]
\centering
\caption{Zero-Shot Classification accuracy on LogiQA Classification with various types of generated reasoning.}
\begin{tabular}{ll}
\hline
Additional Reasoning & Acc. \\
\hline
None (Zero-Shot) & 0.420 \\
Random Reasoning & 0.405 \\
Max ROSCOE reasoning & 0.438 \\
Short-path reasoning (Ours) & \textbf{0.459} \\
\hline
\end{tabular}

\label{tab:logiqa_zeroshot}
\end{table}

Alternatively, we can assess the quality of generated reasoning by adding the generated reasoning during zero-shot prediction. Specifically, we compare three different types of generated reasoning (1) Zero-Shot Reasoning, (2) Max ROSCOE Reasoning, and (3) Short-Path Reasoning (Ours) by appending the chain-of-thought reasoning in-context. We report the results in \Cref{tab:logiqa_zeroshot}. Here, we observe three takeaways. First, merely appending random zero-shot reasoning provides a small benefit over using no reasoning at all ($\uparrow0.012$), suggesting a generic “reasoning-priming” effect. Second, higher-quality reasoning paths help more: Max-ROSCOE improves to $0.571$ accuracy ($\uparrow0.025$). Third, our pivot-focused SPR yields the largest gains ($\uparrow0.053$), outperforming Max-ROSCOE by $\uparrow0.028$ accuracy. Notably, the F1 improvements consistently exceed accuracy gains, indicating that concise, pivot-dense paths improve positive/negative discrimination rather than only shifting the decision threshold. These results support our hypothesis that the transfer signal comes from shared decision pivots rather than from longer or stylistically rich explanations.

\paragraph{Effect of backbone on downstream task.}

\begin{table}[h!]
\centering
\caption{Zero-Shot accuracy on LogiQA with different backbones.}
\begin{tabular}{ll}
\hline
\textbf{Method} & \textbf{Overall Acc.} \\
\hline
Zero-Shot (Qwen-3 0.6B) & 0.420 \\
Zero-Shot (Qwen-3 1.7B) & 0.643 \\
Zero-Shot (Qwen-3 4B) & 0.726 \\
Zero-Shot (Qwen-3 8B) & 0.747 \\
\hline
\end{tabular}
\label{tab:backbone_logia}
\end{table}

In \Cref{tab:backbone_logia}, we analyze the effect of the model on the zero-shot downstream task accuracy. In zero-shot classification, accuracy improves monotonically with model scale from 0.420 (Qwen-3 0.6B) to 0.643 (1.7B), 0.726 (4B), and 0.747 (8B), a net gain of +0.327. The largest jump occurs between 0.6B→1.7B (+0.223), with diminishing returns thereafter (+0.083 from 1.7B→4B; +0.021 from 4B→8B). Consistent with our experimental setup, we use the 0.6B backbone as the default for controlled ablations and self-training loops, and treat larger models as scale references to contextualize improvements.

\paragraph{Comparison of Inference Latency}

\begin{table}[h!]
\centering
\caption{Average Inference Time of varying models.}
\begin{tabular}{l c}
\toprule
\textbf{Method (with Thinking)} & \textbf{Latency per Sample} \\
\midrule
Qwen-3-0.6B & 8--12 s \\
Qwen-3-1.7B & 14--20 s \\
Qwen-3-4B & 25--27 s \\
Qwen-3-8B & 42--48 s \\
Qwen-3-14B & 55--60 s \\
\bottomrule
\end{tabular}
\label{tab:time}
\end{table}

As shown in \Cref{tab:time}, end-to-end latency with chain-of-thought increases with model size—8–12\,s (Qwen-3 0.6B), 14–20\,s (1.7B), 25–27\,s (4B), 42–48\,s (8B), and 55–60\,s (14B) per sample. Because our data-generation pipeline emits $K$ reasoning paths per example (with possible re-sampling), the wall-clock cost grows approximately linearly with $K$. For instance, at $K{=}10$ this corresponds to $\sim$1.3–2.0\,min (0.6B), $\sim$4.2–4.5\,min (4B), $\sim$7–8\,min (8B), and $\sim$9–10\,min (14B) per example. Thus, both model size and $K$ are key efficiency knobs; our short-path synthesis (which yields shorter paths) helps amortize corpus-generation costs at scale.

\paragraph{Zero-shot vs.\ Guided Reasoning for Pivot Diversity}
At a high level, we exploit two sources of generated reasoning to widen coverage of the task’s decision-pivot space. \textit{Zero-shot correct} reasoning paths reveal the model’s canonical, high-probability pivot sets (i.e., what the model already trusts), while \textit{guided} paths (generated after revealing $y$ on initial mispredictions) allow the model to route around its default heuristics and surface counter-heuristic or novel pivots that still justify $y$. Aggregating both sources before short-path synthesis yields a pool of successful paths that agree on the label yet differ in which pivots they invoke and how they organize them. This, in turn, improves the stability of pivot extraction and increases the chance that the final short-path reasoning captures minimal but diverse justifications that generalize.

\subsection{Experimental Details}\label{appendix:setting}

In this section, we report the experimental details of our paper. In all of our experiments, we have used the HuggingFace's $\textsc{trl}$ (Transformer Reinforcement Learning
) library \cite{vonwerra2022trl}. Regarding the datasets mentioned throughout our work, we use three public benchmarks, \textsc{LogiQA} \cite{liu2020logiqa}, \textsc{MedQA} \cite{jin2021disease}, and \textsc{MATH500} \cite{hendrycks2021measuring}, all available through Huggingface's $\textsc{datasets}$ library \cite{lhoest-etal-2021-datasets}. In all datasets, we used the default train-test split provided in the original papers. Regarding the hyperparameters (e.g., SFT learning rate and RL learning rate, number of training epochs), we used the default setting provided in the $\textsc{trl}$ library. For instance, the DPO learning rate was set as $1e-06$, the epochs set as $3$, using LoRA attention dimension of $16$ with alpha of $32$. When performing SFT, we used $2$ epochs with a learning rate of $4e-05$. Regarding unique hyperparameters, we set $K$ (the number of reasoning candidate pool) as $7$ unless otherwise noted and evaluated on a single self-training loop. Please see \Cref{tab:ablation_overview} in \Cref{sec:experiments} for a detailed explanation of the hyperparameter-tuning results. For all experiments, we used the Qwen-3 0.6B model \cite{yang2025qwen3}. Regarding the human-annoated set of 500 Q\&A samples with pivot annotations in \Cref{tab:pivot_retrieval}, we must underline that these annotations are used solely for diagnostic analysis and do not affect training or model selection. Lastly, regarding the computing resources, we used five NVIDIA A100 GPUs for all experiments. For our experiments, we used the 2.2.1 version of Pytorch \citep{pytorch}.


\newpage

\begin{figure}[ht] \centering \includegraphics[width=0.55\textwidth]{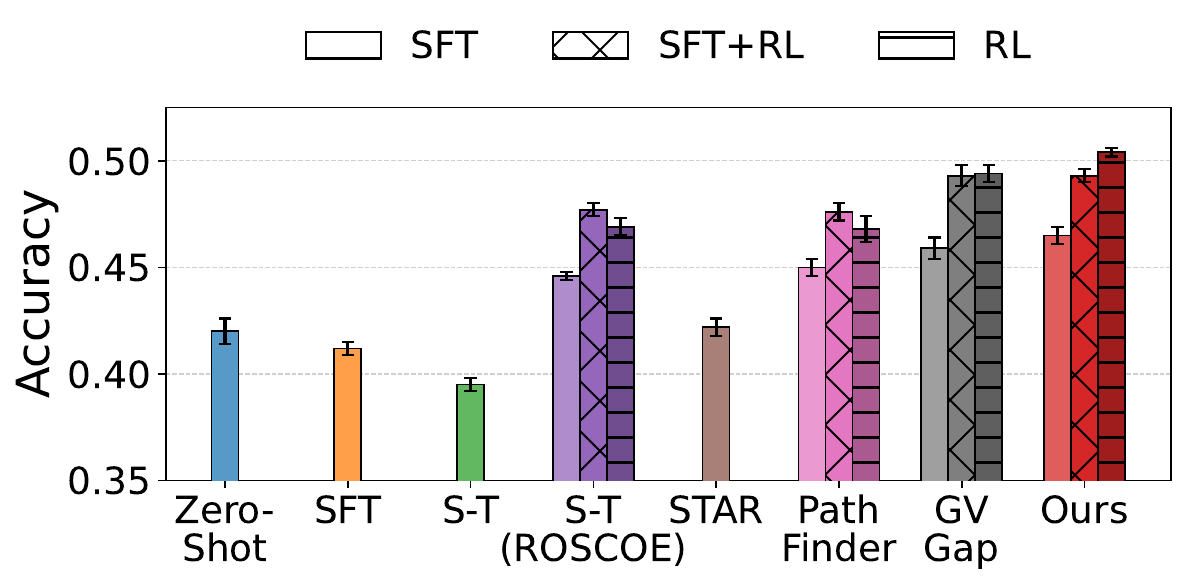} \caption{Comparison of self-training results on LogiQA.} \label{fig:logiqa_solo} \end{figure}

\begin{figure}[ht] \centering \includegraphics[width=0.55\textwidth]{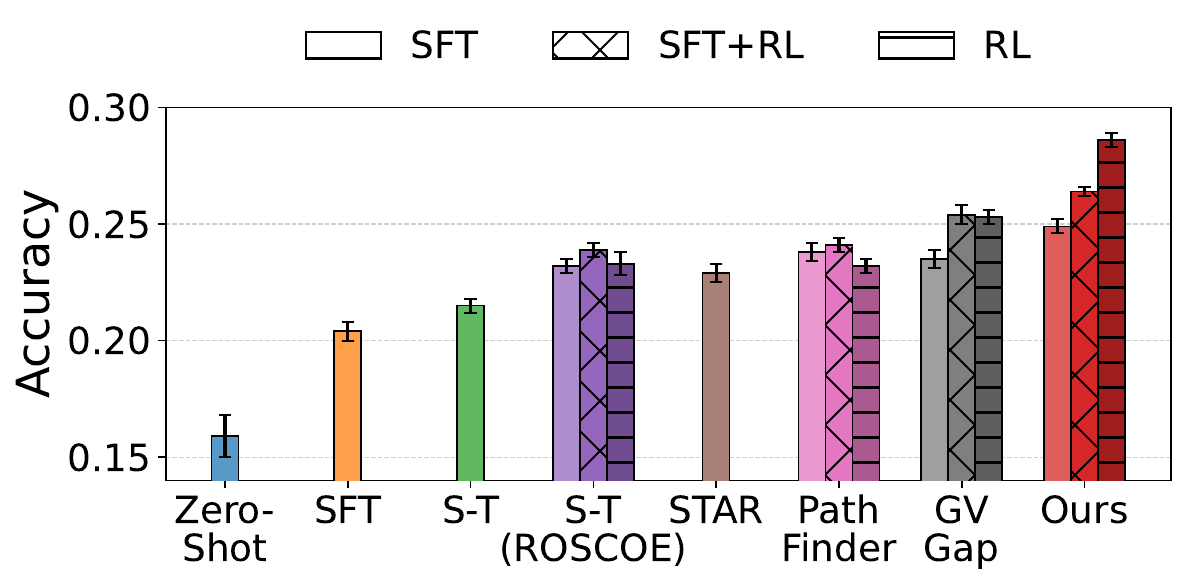} \caption{Comparison of self-training results on MedQA.} \label{fig:medqa_solo} \end{figure}

\begin{figure}[ht] \centering \includegraphics[width=0.55\textwidth]{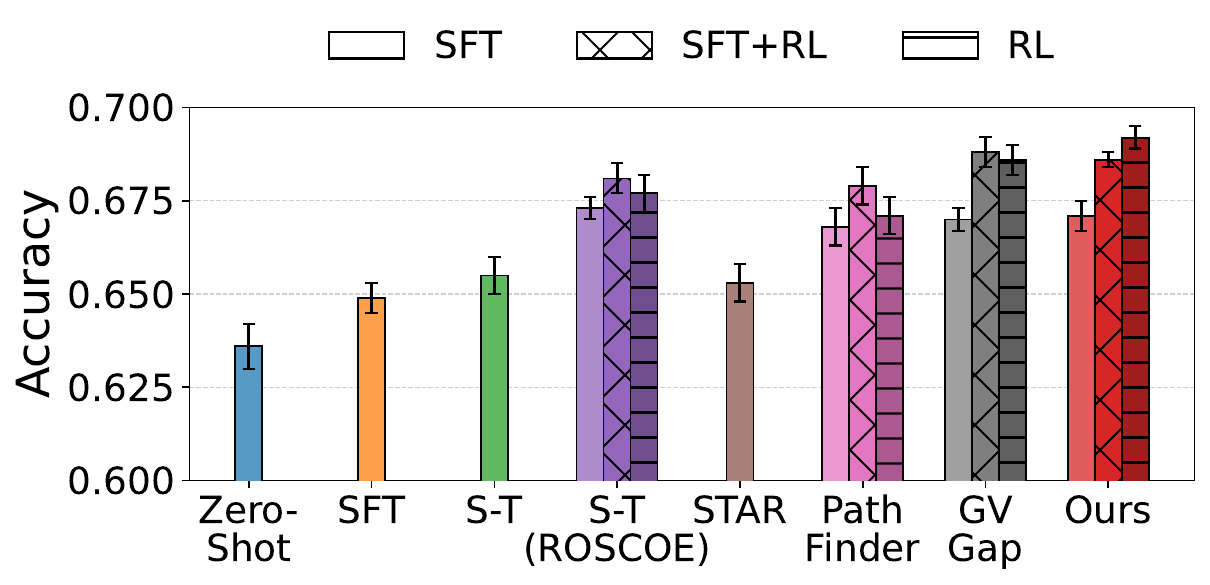} \caption{Comparison of self-training results on MATH500.} \label{fig:math500_solo} \end{figure}

\begin{figure}[t]
\centering
\includegraphics[width=0.6\linewidth]{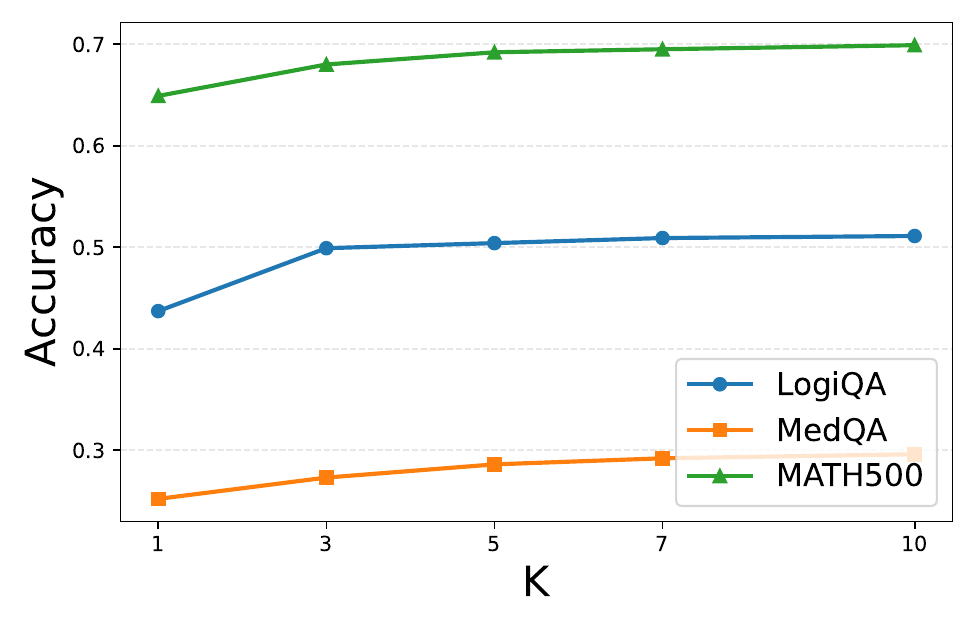}
\caption{Effect of the candidate pool size $K$ on downstream accuracy for LogiQA, MedQA, and MATH500. Larger $K$ steadily improves performance, with diminishing returns near $K{=}10$.}
\label{fig:ablation_k}
\end{figure}

\begin{figure}[t]
\centering
\includegraphics[width=0.6\linewidth]{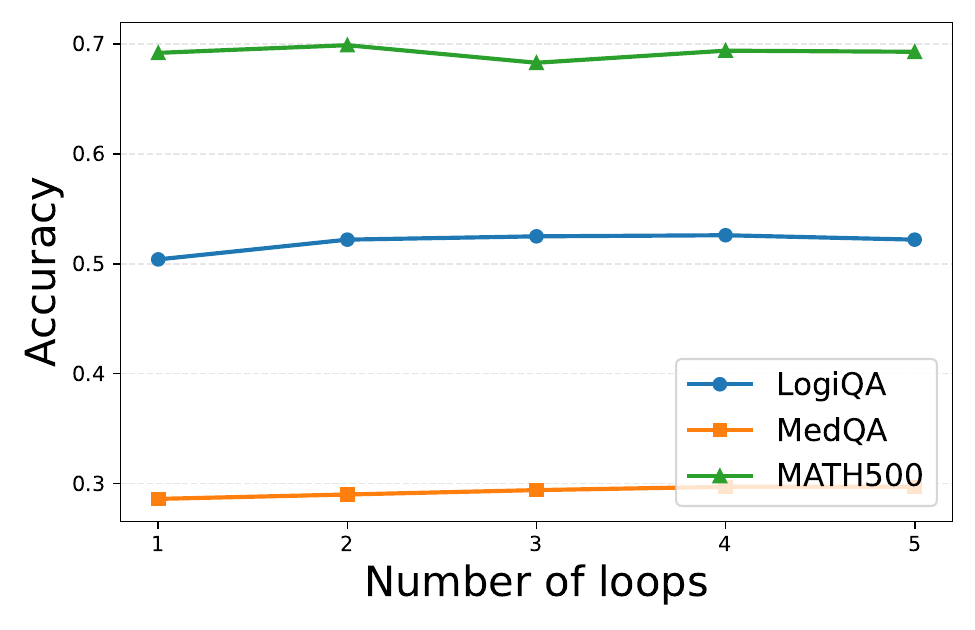}
\caption{Effect of the number of self-training loops on downstream accuracy. Moderate looping (e.g., 2–4) helps, while excessive looping can overfit or oscillate.}
\label{fig:ablation_loops}
\end{figure}

\end{document}